\documentclass[sigconf]{aamas} 
\usepackage{balance} % for balancing columns on the final page
\usepackage{amsmath,amsthm}
\usepackage{mathrsfs}
\usepackage{bm}
\usepackage{graphicx}
\usepackage[ruled,linesnumbered,norelsize,noend]{algorithm2e}
\usepackage{multirow}
\usepackage{booktabs} % for professional tables
\usepackage{csquotes}
\usepackage{subfigure}

\setcopyright{ifaamas}
\acmConference[AAMAS '24]{Proc.\@ of the 23rd International Conference
on Autonomous Agents and Multiagent Systems (AAMAS 2024)}{May 6 -- 10, 2024}
{Auckland, New Zealand}{N.~Alechina, V.~Dignum, M.~Dastani, J.S.~Sichman (eds.)}
\copyrightyear{2024}
\acmYear{2024}
\acmDOI{}
\acmPrice{}
\acmISBN{}

\acmSubmissionID{<<EasyChair submission id>>}

\title[Grasper: A Generalist Pursuer for Pursuit-Evasion Problems]{Grasper: A Generalist Pursuer for Pursuit-Evasion Problems}

\author{Pengdeng Li$^{\ast}$}\thanks{$^{\ast}$Equal contribution.}
\affiliation{
  \institution{Nanyang Technological University}
  \city{Singapore}
  \country{}}
\email{pengdeng.li@ntu.edu.sg}

\author{Shuxin Li$^{\ast}$}
\affiliation{
  \institution{Nanyang Technological University}
  \city{Singapore}
  \country{}}
\email{shuxin.li@ntu.edu.sg}

\author{Xinrun Wang$^{\dagger}$}\thanks{$^{\dagger}$Corresponding author.}
\affiliation{
  \institution{Nanyang Technological University}
  \city{Singapore}
  \country{}}
\email{xinrun.wang@ntu.edu.sg}

\author{Jakub \v{C}ern\'{y}}
\affiliation{
  \institution{Columbia University}
  \city{New York City}
  \country{United States}}
\email{cerny@disroot.org}

\author{Youzhi Zhang}
\affiliation{
  \institution{CAIR, HKISI, CAS}
  \city{Hong Kong}
  \country{China}}
\email{youzhi.zhang@cair-cas.org.hk}

\author{Stephen McAleer}
\affiliation{
  \institution{Carnegie Mellon University}
  \city{Pittsburgh}
  \country{United States}}
\email{mcaleer.stephen@gmail.com}

\author{Hau Chan}
\affiliation{
  \institution{University of Nebraska-Lincoln}
  \city{Lincoln, Nebraska}
  \country{United States}}
\email{hchan3@unl.edu}

\author{Bo An}
\affiliation{
  \institution{Nanyang Technological University}
  \city{Singapore}
  \country{}}
\email{boan@ntu.edu.sg}

\begin{abstract}
Pursuit-evasion games (PEGs) model interactions between a team of pursuers and an evader in graph-based environments such as urban street networks. Recent advancements have demonstrated the effectiveness of the pre-training and fine-tuning paradigm in Policy-Space Response Oracles (PSRO) to improve scalability in solving large-scale PEGs. However, these methods primarily focus on specific PEGs with fixed initial conditions that may vary substantially in real-world scenarios, which significantly hinders the applicability of the traditional methods. To address this issue, we introduce Grasper, a GeneRAlist purSuer for Pursuit-Evasion pRoblems, capable of efficiently generating pursuer policies tailored to specific PEGs. Our contributions are threefold: First, we present a novel architecture that offers high-quality solutions for diverse PEGs, comprising critical components such as (i) a graph neural network (GNN) to encode PEGs into hidden vectors, and (ii) a hypernetwork to generate pursuer policies based on these hidden vectors. As a second contribution, we develop an efficient three-stage training method involving (i) a pre-pretraining stage for learning robust PEG representations through self-supervised graph learning techniques like graph masked auto-encoder (GraphMAE), (ii) a pre-training stage utilizing heuristic-guided multi-task pre-training (HMP) where heuristic-derived reference policies (e.g., through Dijkstra's algorithm) regularize pursuer policies, and (iii) a fine-tuning stage that employs PSRO to generate pursuer policies on designated PEGs. Finally, we perform extensive experiments on synthetic and real-world maps, showcasing Grasper's significant superiority over baselines in terms of solution quality and generalizability. We demonstrate that Grasper provides a versatile approach for solving pursuit-evasion problems across a broad range of scenarios, enabling practical deployment in real-world situations.
\end{abstract}


\newcommand{\BibTeX}{\rm B\kern-.05em{\sc i\kern-.025em b}\kern-.08em\TeX}

\makeatletter
\gdef\@copyrightpermission{
	\begin{minipage}{0.3\columnwidth}
		\href{https://creativecommons.org/licenses/by/4.0/}{\includegraphics[width=0.90\textwidth]{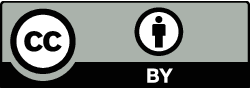}}
	\end{minipage}\hfill
	\begin{minipage}{0.7\columnwidth}
		\href{https://creativecommons.org/licenses/by/4.0/}{This work is licensed under a Creative Commons Attribution International 4.0 License.}
	\end{minipage}
	\vspace{5pt}
}
\makeatother

\begin{document}

%%% The following commands remove the headers in your paper. For final 
%%% papers, these will be inserted during the pagination process.

\pagestyle{fancy}
\fancyhead{}

%%% The next command prints the information defined in the preamble.

\maketitle 

%%%%%%%%%%%%%%%%%%%%%%%%%%%%%%%%%%%%%%%%%%%%%%%%%%%%%%%%%%%%%%%%%%%%%%%%

\section{Introduction}

The deployment of security resources to detect, deter, and catch criminals is a critical task in urban security~\cite{tsai2010urban,tambe2011security}. Statistics show that police pursuits probably \enquote{injure or kill more innocent bystanders than any other kind of force}~\cite{rivara2004motor}. Therefore, it is crucial to come up with scalable approaches for effectively coordinating various security resources, ensuring the swift apprehension of a fleeing criminal to minimize harm and property damage. Due to the adversarial nature between attackers and defenders, game-theoretic models have been used to model various real-world urban security scenarios. In particular, the pursuit-evasion game (PEG) has been extensively employed to model the interactions between a team of pursuers (e.g., police forces) and an evader (e.g., a criminal) on graphs (e.g., urban street networks)~\cite{zhang2017optimal,zhang2019optimal,li2021cfrmix,xue2021solving}. To effectively solve PEGs under various settings, several methods, such as counterfactual regret minimization (CFR) \cite{zinkevich2008regret} and policy-space response oracles (PSRO) \cite{lanctot2017unified}, have been developed in the literature. Among these algorithms, PSRO, a deep reinforcement learning algorithm, provides a versatile framework for learning the (approximate) Nash equilibria (NEs) of PEGs (refer to Section~\ref{sec:psro} for an introduction of the PSRO framework). Furthermore, recent works have also integrated the pre-training and fine-tuning paradigm into the PSRO framework to further enhance its scalability~\cite{li2023solving}.

Although many existing works have achieved significant success, they only focus on solving specific PEGs with predetermined initial conditions, e.g., the initial locations of all players and exits, and the time horizon of the game. Unfortunately, these conditions may vary substantially in real-world scenarios, where crimes can occur at any location in a city and at any time. When the initial conditions change, existing algorithms must solve the new PEG from scratch, which is computationally demanding and time-consuming~\cite{li2023solving}, restricting the real-world deployment of current algorithms. Thus, there is an urgent necessity to develop a new approach capable of solving different PEGs with varying initial conditions effectively.

To this end, we introduce Grasper: a GeneRAlist purSuer for Pursuit-Evasion pRoblems, which can effectively solve different PEGs by generating the pursuer's policies conditional on the initial conditions of the PEGs. Grasper consists of two critical components. First, as the PEG is played on a graph, it is natural to use a graph neural network (GNN) to encode the PEG with the given initial conditions into a hidden vector. Then, inspired by recent work on generalization over games with different population sizes~\cite{li2023populationsizeaware}, we introduce a hypernetwork to generate the base policy for the pursuer conditional on the hidden vector obtained by the GNN. This generated base policy then serves as a starting point for the pursuer's best response policy training at each PSRO iteration. 

To train the networks of Grasper, we find that naively applying multi-task training~\cite{zhang2021survey} is inefficient. Furthermore, jointly training the GNN and hypernetwork could be time-consuming as the GNN is only used to encode the initial conditions which are fixed during the game playing. To address these challenges, we propose an efficient three-stage training method to train the networks of Grasper. First, we introduce a pre-pretraining stage to train the GNN by using self-supervised graph learning methods such as GraphMAE~\cite{hou2022graphmae}. Second, we fix the GNN and pre-train the hypernetwork by using a multi-task training procedure where the training data is sampled from different PEGs with different initial conditions. In this stage, to overcome the low exploration efficiency due to the pursuers' random exploration and the evader's rationality, we propose a heuristic-guided multi-task pre-training (HMP) where a reference policy derived by heuristic methods such as Dijkstra is used to regularize the pursuer policy. Finally, we follow the PSRO procedure and obtain the pursuer's best response policy at each iteration by fine-tuning the base policy generated by the hypernetwork.

In summary, we provide three contributions. First, we propose Grasper which is the first generalizable framework capable of efficiently providing highly qualified solutions for different PEGs with different initial conditions. Second, to efficiently train the networks of Grasper, we propose a three-stage training method: (i) a pre-pretraining stage to train the GNN through GraphMAE, (ii) a pre-training stage to train the hypernetwork through heuristic-guided multi-task pre-training (HMP), and (iii) a fine-tuning stage to obtain the pursuer's best response policy at each PSRO iteration. Finally, we perform extensive experiments, and the results demonstrate the superiority of Grasper over different baselines.

\section{Related Work}

Pursuit-evasion games (PEGs) have been extensively applied to model various real-world problems such as security and robotics~\cite{vidal2002probabilistic,bopardikar2008discrete,horak2017dynamic,lopez2019solutions,wang2020cooperative,huang2021dynamic,li2022survey}. To efficiently solve PEGs and different variants, many algorithms such as value iteration~\cite{horak2017dynamic} and incremental strategy generation~\cite{zhang2017optimal,zhang2019optimal} have been introduced. Nonetheless, these methods encounter scalability issues as they typically rely on linear programming. On the other hand, PEG can be viewed as a particular type of two-player imperfect-information extensive-form game (IIEFG). Thus, the algorithms used for solving large-scale IIEFGs, such as counterfactual regret minimization (CFR)~\cite{zinkevich2008regret} and Policy-Space Response Oracles (PSRO)~\cite{lanctot2017unified}, have been applied to tackle large-scale PEGs~\cite{li2021cfrmix,xue2021solving}. However, when solving large-scale PEGs using PSRO, there exist significant computational challenges as it involves computing the best response strategy multiple times. To mitigate this issue, recent research~\cite{li2023solving} integrates the pre-training and fine-tuning paradigm into PSRO to improve its scalability. Despite the success, all these algorithms are tailored to solve a specific PEG with predetermined initial conditions. When these conditions change, they must recompute the NE strategy from scratch (one to two hours for a PEG on a $10\times10$ grid map~\cite{li2023solving}), which hinders their real-world applicability\footnote{Note that classical heuristic algorithms such as Dijkstra are also less applicable owing to this reason. Moreover, it is less meaningful to assume that there is at least one pursuer at each exit as the pursuer's resources are typically limited~\cite{tsai2010urban}.}. To address this limitation, we propose Grasper, which uses PSRO to compute the NE strategy and can generate different pursuers' strategies for different PEGs based on their initial conditions without recomputing the NE strategy from scratch. 

The generalizability of algorithms and models over different games has gained increasing attention and remarkable progress has been achieved in recent research. Neural equilibrium approximators that directly predict the equilibrium strategy from game payoffs in normal-form games (NFGs) have been theoretically proven PAC learnable~\cite{duan2023nash,duan2023equivariant} and are able to generalize to different games with desirable solution quality~\cite{feng2021neural,marris2022turbocharging,duan2023nash,duan2023equivariant}. However, it remains under-explored when going beyond NFGs. In this work, we make the first attempt to consider the generalization problem in the domain of PEGs, a type of game that has a wide range of real-world applications~\cite{horak2017dynamic,li2022survey} and is far more complicated than NFGs. We propose a novel algorithmic framework that is able to efficiently solve different PEGs with varying initial conditions and demonstrate the generalization ability through extensive experiments.

Our work is also related to self-supervised graph learning and multi-task learning. Recent works have shown that generative self-supervised learning~\cite{he2022masked} can be applied to graph learning and outperform contrastive methods which require complex training strategies~\cite{qiu2020gcc,thakoor2022largescale}, high-quality data augmentation~\cite{zhang2021canonical}, and negative samples that are often challenging to construct from graphs~\cite{zhu2021graph}. Therefore, we employ the recent state-of-the-art, GraphMAE~\cite{hou2022graphmae}, to learn a good representation of a PEG with the given initial conditions. Multi-task learning~\cite{ruder2017overview,zhang2021survey} has been applied to various domains including natural language processing~\cite{collobert2008unified}, computer vision~\cite{girshick2015fast}, and reinforcement learning (multi-task RL)~\cite{wilson2007multi,vuong2019sharing,zeng2021decentralized,mandi2022effectiveness}. Due to its strong generalizability, we employ multi-task RL for the pre-training process, enabling the pre-trained policy to be quickly fine-tuned for efficient policy development in new tasks.

Finally, our work is related to the multi-agent patrolling problem where the evader often anticipates patrolling strategies and may choose a target or a single path as an action~\cite{agmon2011multi,sless2014multi,huang2019survey,buermann2022multi}. Conversely, our pursuit-evasion game features simultaneous actions with the evader unaware of the pursuer’s real-time locations.

\section{Problem Formulation\label{sec:problem_formulation}}

In this section, we first present all the elements for defining the PEGs. Then, we present the state-of-the-art (SOTA) method for solving PEGs. Finally, we give the problem statement of this work.

\subsection{Preliminaries\label{sec:preliminaries}}

A pursuit-evasion game (PEG) is a two-player game played between a pursuer and an evader, i.e., $N=\{p, e\}$. Following previous works \cite{zhang2019optimal,li2021cfrmix,xue2021solving}, we assume that the pursuer comprises $n$ members denoted as $p= \{1, 2, ..., n\}$, and the pursuer can obtain the real-time location information of the evader with the help of tracking devices. In reality, PEGs are typically played on urban road maps, which can be represented by a graph $G=(V, E)$, where $V$ is the set of vertices and $E$ is the set of edges. Let $V^{\prime} \subset V$ denote the set of exit nodes from which the evader can escape and $T$ the predetermined time horizon of the game. At $t \leq T$, the locations of the evader and pursuer are denoted by $l_t^e$ and $l_t^{p}=(l_t^1, l_t^2, ..., l_t^n)$, respectively. Then, the history of the game at $t$ is a sequence of past locations of both players, i.e., $h = (l_0^e, l_0^{p}, ..., l_{t-1}^e, l_{t-1}^{p})$. The available action set for both players is the neighboring vertices of the player's current location, i.e., $A_e(h)=\mathcal{N}(l^e_{t-1})$ and $A_{p}(h) = \{(l^1, l^2, ..., l^n)|l^i \in \mathcal{N}(l^i_{t-1}), \forall i \in p\}$ where $\mathcal{N}(v)$ denotes the set of neighboring vertices of vertex $v$. According to the definition of history, we define the information set for each player as the set consisting of indistinguishable histories. As the evader cannot get the pursuer's real-time location information, the information set of the evader is defined as $I_e = \{h|h=(l^e_0, l^{p}_0, l^e_1, *, ..., l^e_{t-1}, *)\}$, where $*$ represents any possible location of the pursuer. Although the PEG is a simultaneous-move game, we can model it as an extensive-form game (EFG) by assuming that the evader acts first and then the pursuer commits without any information about the evader's current action. The pursuer's information set can be defined as $I_{p}=\{h|h=(l^e_0, l^{p}_0, ..., l^e_{t-1}, l^{p}_{t-1}, *)\}$ since the pursuer knows the evader's location but not the evader's current action. 

A behavior policy of a player assigns a probability distribution over the action set for every information set belonging to the player. Notice that the pursuer's action space is combinatorial and expands exponentially with the number of pursuer members. As a result, directly learning a joint policy of the pursuer members would be computationally difficult. To address this issue, instead of learning a joint policy, previous works learn the individual policies either through value decomposition~\cite{li2021cfrmix} or global critic~\cite{li2023solving}, which are the paradigm of centralized training with decentralized execution (CTDE) for the pursuers. Furthermore, previous works~\cite{li2023solving} also introduce a new state representation ignoring the game's historical information, which leads to improved performance. In our work, we follow the previously mentioned conventions to define the observations and individual policies for the pursuers. 

At each time step $t$, each pursuer member gets an \textbf{observation} $o_t^i=(l_t^{p}, l_t^e, i, t) \in \mathcal{O}^i$, which includes all players' current locations, the id of the pursuer member, and the time step. Each pursuer member $i$ constructs a \textbf{policy}\footnote{All the pursuer members share one policy. As the observations include pursuers' ids, different pursuer members can have different behaviors~\cite{foerster2018counterfactual}. $\Delta$ denotes the simplex.} $\pi^p: \mathcal{O}^i \to \Delta(A_i)$, which assigns a probability distribution over the \textbf{action set} $A_i(o_t^i)=\mathcal{N}(l_t^i)$, $\forall i \in p$. As for the evader's policy, we follow previous works~\cite{xue2021solving,xue2022nsgzero} that employ High-Level Actions for the evader. That is, the evader only chooses the exit node to escape from and then samples one shortest path from the initial location to the chosen exit node, instead of deciding where to go in the next time step. Specifically, at time step $t=0$, the evader determines an exit node $v^{\prime} \in V^{\prime}$ using the \textbf{policy} $\pi^e: V \to \Delta(V^{\prime})$, samples a shortest path from $l_0^e \in V$ to the chosen exit node $v^{\prime}$, and then takes actions based on the path. 

Here, we give some remarks on the assumption of High-Level Actions of the evader. (i) In our game setting, the evader lacks real-time access to the pursuers' locations, requiring the evader to act without any information about their whereabouts. Therefore, sampling one path for the evader would not lose much information compared with the case where the evader acts step by step. (ii) Training the pursuer against an evader who chooses the shortest path, a worst-case scenario for the pursuer, enhances the robustness of the pursuer's policy. (iii) Though it is a simplification, the problem setting remains highly complex due to the multiple exits and diverse players’ initial conditions, enlarging the task space for the pursuer, as detailed in the Introduction and Appendix A Q2.

In summary, given a graph $G$ with the specific set of exit nodes $V^{\prime}$, the initial locations of the pursuer and evader ($l_0^{p}, l_0^{e}$), and the predetermined time horizon $T$, we can define a specific PEG as $\mathcal{G}=(G, V^{\prime}, l_0^p, l_0^{e}, T)$. In the PEG, players will get the non-zero rewards only when the game is terminated. The termination conditions include three cases: (i) the pursuer catches the evader within the time horizon $T$, i.e., $l_t^e \in l_t^{p}, t \leq T$; (ii) the evader escapes from an exit node within the time horizon $T$, i.e., $l_t^e \in V^{\prime}, t \leq T$; (iii) the game reaches the time horizon $T$. Let $t^{\prime}\leq T$ be the time step that the game is terminated. Then, for all $t<t^{\prime}$, $r_t^{p}=r_t^{e}=0$. For $t=t^{\prime}$, in cases (i) and (iii), the pursuer receives a \textbf{reward} $r_t^{p}=1$ while the evader incurs a penalty $r_t^e=-1$. In case (ii), the evader gains a reward $r_t^e=1$, and the pursuer suffers a loss $r_t^{p}=-1$. Given the exit node chosen by the evader $v^{\prime}\sim\pi^e$, we have $V^p(\pi^p, v^{\prime}) = \mathbb{E}\big[\sum_{t=0}^{T} r_t^p\big]$ for the pursuer and $V^e(\pi^p, v^{\prime}) = \mathbb{E}\big[\sum_{t=0}^{T} r_t^e\big]$ for the evader, where the expectation is taken over the trajectories induced by $\pi^p$. Then, for the policy pair $(\pi^p, \pi^e)$, we have $V^p(\pi^p, \pi^e) = \mathbb{E}_{v^{\prime}\sim\pi^e}\big[V^p(\pi^p, v^{\prime})\big]$ for the pursuer and $V^e(\pi^p, \pi^e) = \mathbb{E}_{v^{\prime}\sim\pi^e}\big[V^e(\pi^p, v^{\prime})\big]$ for the evader.

\subsection{Policy-Space Response Oracles\label{sec:psro}}

\begin{algorithm}[ht]
    \caption{PSRO for a specific PEG $\mathcal{G}$}
    \label{alg:psro}
    $\Pi_0^p=\{\pi_0^p\}$, $\Pi_0^e=\{\pi_0^e\}$, $U_0$, $\sigma_0^p$, $\sigma_0^e$\;
    \For{$\text{epoch k} = 1, 2, \cdots K$}{
        Compute the evader's BR policy $\pi_k^e$ against $\sigma_{k-1}^p$\;
        Compute the pursuer's BR policy $\pi_k^p$ against $\sigma_{k-1}^e$\;
        Expansion: $\Pi_k^p = \Pi_{k-1}^p \cup \{\pi_k^p\}$, $\Pi_k^e = \Pi_{k-1}^e \cup \{\pi_k^e\}$\;
        Update meta-game matrix $U_k$ through simulation\;
        Compute $\sigma_k^p$ and $\sigma_k^e$ using a meta-solver on $U_k$\;
    }
    \textbf{Return}: $\Pi_K^p$, $\Pi_K^e$, $\sigma_K^p$, $\sigma_K^e$
\end{algorithm}

\begin{figure*}[ht]
    \centering
    \includegraphics[width=\textwidth]{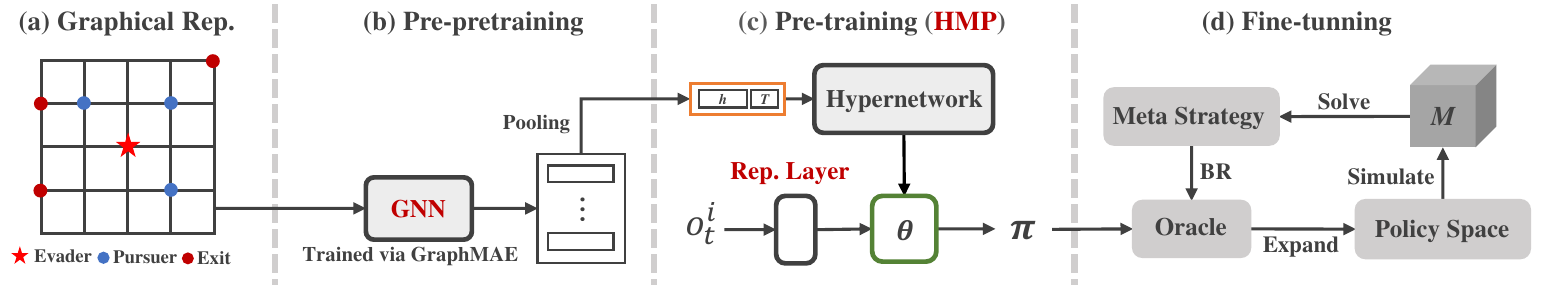}
    \caption{Architecture and training pipeline of Grasper.}
    \label{fig:train_stages}
    \Description{This figure graphically shows the architecture and training pipeline of Grasper.}
\end{figure*}

As one of the popular algorithms, PSRO~\cite{lanctot2017unified} can be employed to solve a PEG $\mathcal{G}$, shown in Algorithm~\ref{alg:psro}. It commences with each player using a random policy (Line 1) and then expands the policy spaces of the pursuer and evader in an iterative manner. At each epoch $1\leq k\leq K$: (1) Compute the best response (BR) policies of the pursuer $\pi_k^p$ and evader $\pi_k^e$ and add them to their policy spaces $\Pi_k^p$ and $\Pi_k^e$ (Line 3--5); (2) Construct a meta-game $U_k$ using all policies in each player's policy space (Line 6); (3) Compute the meta-strategy of the pursuer $\sigma_k^p \in \Delta(\Pi_k^p)$ and evader $\sigma_k^e \in \Delta(\Pi_k^e)$ using a meta-solver (e.g., PRD~\cite{lanctot2017unified}) on the meta-game $U_k$ (Line 7). These processes are repeated for $K$ epochs and then output the final meta-strategy across the players' policy spaces (Line 9).

As the evader’s policy is a probability distribution over exit nodes, to compute the BR policy (Line 3), we only need to estimate the value of each exit node through simulations, i.e., $V^e(v^{\prime}) = \mathbb{E}_{\pi^p\sim \sigma_{k-1}^p}\big[V^e(\pi^p, v^{\prime})\big]$, $\forall v^{\prime} \in V^{\prime}$. Then, the evader's BR policy is constructed by applying softmax operation on the values of all the exit nodes. For the pursuer, computing the BR policy is to solve the problem $\pi_k^p=\arg\max_{\pi^p} \mathbb{E}_{\pi^e\sim\sigma_{k-1}^e}\big[V^p(\pi^p, \pi^e)\big]$ (Line 4). As there are multiple pursuer members, we can use MAPPO~\cite{yu2022surprising} to learn the BR policy. In the traditional PSRO algorithm, the pursuer's BR policy is learned from scratch, i.e., the BR policy is randomly initialized, which is inefficient. To address this issue, recent works integrate the pre-training and fine-tuning paradigm into PSRO to improve learning efficiency~\cite{li2023solving}. Specifically, before running PSRO, a base pursuer policy is trained through multi-task RL where each task is generated with a randomly initialized evader's policy. Then, at each PSRO epoch, the pursuer's BR policy is initialized with the pre-trained base policy, rather than learning from scratch, which can largely improve the learning efficiency of the PSRO algorithm.

\subsection{Problem Statement}
 
Although PSRO has been successfully applied to solve PEGs, unfortunately, previous works typically focus on solving a specific PEG with predetermined initial conditions which are not always fixed in real-world scenarios: (i) The initial locations of the pursuers and the evader ($l_0^{p}, l_0^{e}$) are not always fixed since attacks (thieves, crimes, terrorists) can occur at any time and location in a city; (ii) The locations of the exit nodes $V^{\prime}$ may change due to temporary closures and openings; (iii) The time horizon $T$ might vary, as the time required to pursue the evader is not always the same. When any of the initial conditions change, the PEG adapts accordingly. As a consequence, current algorithms can only solve the modified PEG from scratch, leading to significant time consumption and inefficiency. Even the SOTA method presented in the previous section -- PSRO with pre-trained base pursuer policy -- still suffers from such an issue as the base policy is pre-trained under the premise that the initial condition of the PEG is fixed. In other words, a new base policy must be pre-trained from scratch for the modified PEG since the original base policy may not be a good starting point for the pursuer's BR policy in the modified game (even worse than a randomly initialized BR policy). In this paper, we aim to address this issue by developing a generalist pursuer capable of learning and adapting to different PEGs with varying initial conditions without the need to restart the training process from the beginning.

\section{Grasper}

In this section, we introduce Grasper, illustrated in Figure~\ref{fig:train_stages}. We first present the architecture of Grasper including several innovative components, and then the training pipeline which consists of three stages to efficiently train the networks of Grasper.

\subsection{Architecture}

\subsubsection{\textbf{Graphical Representations of PEGs.}} To generate the pursuer's policy based on the specific PEG, we propose to take the specific PEG as an input of a neural network. To this end, we encode the initial conditions of a PEG except for $T$ into a graph (Figure~\ref{fig:train_stages}(a)). The time horizon $T$ can be directly fed into the neural network. Specifically, given a PEG $\mathcal{G}=(G, V^{\prime}, l_0^p, l_0^{e}, T)$, these initial conditions $V^{\prime}, l_0^p$ and $l_0^{e}$ can be encoded into the graph $G$ by associating each node of the graph with a vector consisting of the following parts: (i) a binary bit $\{0, 1\}$ where $1$ indicates that the node is an exit, (ii) a binary bit $\{0,1\}$ where $1$ signifies that the evader's initial location is this node, (iii) the number of pursuers on this node $\{0, \dots, n\}$ (the total number of pursuers across all nodes equals to $n$), and (iv) additional information regarding the topology of the graph, such as the degree of the node. This provides a universal representation of any PEG with any initial condition.

\subsubsection{\textbf{Game-conditional Base Policies Generation.}} After representing a PEG as a graph, it is natural to leverage a graph neural network (GNN) to encode the PEG with the given initial conditions into a hidden vector. As shown in Figure~\ref{fig:train_stages}(b), we first feed the graphical representation of the initial conditions into the GNN and get the representations of all the nodes of the graph. Then, we use a pooling operation to integrate all the node representations into a hidden vector which will be concatenated with the time horizon $T$ to get the final representation of the PEG. Next, to generate a base policy conditional on the PEG, we introduce a hypernetwork~\cite{ha2017hypernetworks}, a neural network that takes the final representation of the PEG as input and outputs the parameters (weights and biases) of the policy network (Figure~\ref{fig:train_stages}(c)). Finally, the base policy network serves as a starting point for the training of the pursuer's best response policy in each iteration of the PSRO algorithm (Figure~\ref{fig:train_stages}(d)).

\subsubsection{\textbf{Observation Representation Layer.}} As described earlier, the pursuer's policy is a mapping that associates each observation with a probability distribution over the available action set. Notably, an observation consists of the positions of both players. Representing these positions by mere index numbers of vertices in the graph does not provide much useful information for training, though. Therefore, we seek a more compact and meaningful representation of these observations. Previous works~\cite{xue2021solving,li2023solving} typically train a node embedding model for this purpose. Unfortunately, such a model is often tailored and trained for a specific graph, limiting its generalizability to other graphs. This lack of generalizability makes this method unsuitable for our problem. To address this issue, we adopt a representation layer to encode the pursuer's observations, an approach that is not limited to a specific graph. 

As given in Section~\ref{sec:preliminaries}, the pursuer's observation $o_t^i=(l_t^{p}, l_t^e, i, t)$ includes three parts: the players' current locations $(l_t^{p}, l_t^e)$, the pursuer member's id $i$, and the current time step $t$. Thus, the representation layer consists of three components, each of which is a \enquote{torch.nn.Embedding} which has been extensively used to encode an integer to a compact representation. The outputs of the three components are concatenated to obtain the representation of the pursuer’s observation. This representation layer will be trained jointly with the hypernetwork during the pre-training process. Please refer to Appendix B for details on the architecture of the representation layer, the GNN, and the hypernetwork.

Intuitively, the generalization ability of Grasper benefits from several designs of our architecture. First, the graphical representation offers a universal representation of any PEG, regardless of the graph's topology. Second, GNN can encode different PEGs into fixed-size hidden vectors, which can be directly fed into the hypernetwork (otherwise, additional techniques are required if the sizes of the hidden vectors are varied). Finally, the hypernetwork is designed to generate a specialized policy tailored to a given PEG. 

\subsection{Training Pipeline}

Now we introduce the training pipeline of Grasper, which involves three stages: pre-pretraining, pre-training, and fine-tuning. Prior to delving into the specifics, we first describe how the training set is generated. The training set $\mathcal{I}$ should consist of different PEGs for training. To this end, we generate the training set by randomizing the initial conditions, denoted by $(G, V^{\prime}, l_0^p, l_0^{e}, T)$. However, this approach may yield certain games that lack meaningful training value. For example, when the evader's initial location is in such close proximity to the exit nodes that the pursuer becomes incapable of capturing the evader regardless of its movements. To exclude these trivial cases, we introduce a filter condition when generating the training set: the shortest path from the evader's initial location to any exit nodes must exceed a predetermined length. 
 
\subsubsection{\textbf{Stage I: Pre-pretraining.}} As the hypernetwork takes a feature vector as input, we first use a GNN to encode the graphical representation of the PEG into a fixed-size hidden vector. As the GNN is solely employed to derive the effective representation from the PEG's graphical interpretation, we introduce a pre-pretraining stage (Figure~\ref{fig:train_stages}(b)) to pre-train the GNN before the actual pre-training stage. This approach is more efficient compared to jointly training the GNN and hypernetwork in the pre-training stage. Specifically, for each game in the training set $\mathcal{G} \in \mathcal{I}$, let $\bm{A}_{\mathcal{G}}$ and $\bm{X}_{\mathcal{G}}$ denote the adjacency matrix and feature matrix of the underlying graph, respectively. We first obtain the latent code matrix $\bm{H}_{\mathcal{G}} = f_{\text{GNN}}(\bm{X}_{\mathcal{G}}, \bm{A}_{\mathcal{G}})$ by the GNN and train the GNN via any self-supervised graph learning method (we use the recent SOTA method, GraphMAE~\cite{hou2022graphmae}). Then, we get the hidden vector by pooling the latent code matrix $\bm{h}_{\mathcal{G}} = pool(\bm{H}_{\mathcal{G}})$, which will be fed into the hypernetwork. 

\begin{algorithm}
    \caption{Pre-training}
    \label{alg:pre-training}
    Initialize Grasper and the episode buffer $\mathcal{D} \gets \emptyset$\;
    \For{$\text{train epoch} = 1, 2, \cdots$}{
        \textcolor{blue}{Uniformly sample $c_1$ games from the training set $\mathcal{I}$}\;
        \For{\textcolor{blue}{each of the $c_1$ games $\mathcal{G}$}}{
            Randomly generate $c_2$ evader's policies\;
            \textcolor{blue}{Generate pursuer's policy $\pi^p_{\bm{\theta}} \gets \text{Grasper}(\mathcal{G})$}\;
            \For{each of the $c_2$ evader's policies $\pi^e$}{
                \textcolor{blue}{Sample data using $\pi^e$, $\pi^p_{\bm{\theta}}$, and $\hat{\pi}^p$}\;
                Add the data into the episode buffer $\mathcal{D}$\;
            }
        }
        Train the networks by optimizing the \textcolor{blue}{loss function $L$}\;
        Clear the episode buffer $\mathcal{D} \gets \emptyset$\;
    }
\end{algorithm}

\subsubsection{\textbf{Stage II: Pre-training.}} Given a fixed evader's policy in a specific PEG, computing the pursuer's best response policy can be seen as an RL task. Thus, we can apply the multi-task RL algorithm to guide the pre-training process, which is shown in Algorithm~\ref{alg:pre-training}. Different from previous work~\cite{li2023solving}, in these RL tasks, except for the change in the evader's policy, the game's initial conditions also change. To obtain these RL tasks for pre-training, we first randomly sample $c_1$ games from the training set (Line 3), and then for each game, we randomly sample $c_2$ evader's policies (Line 5). Once the game and the evader's policy are fixed, the RL task is generated. During pre-training, for each game, we first feed the hidden vector of the game (obtained by the trained GNN) and the time horizon into the hypernetwork to generate the pursuer's base policy, and then for each evader's policy, we collect the training data using the pursuer's base policy (accompany by the representation layer) into the episode buffer. Finally, we train the hypernetwork and the representation layer jointly based on the episode buffer (Lines 6-9). To deal with the multiple pursuer members cases, we employ MAPPO~\cite{yu2022surprising} as the underlying RL algorithm.

\begin{figure}[ht]
\centering
\includegraphics[width=0.67\columnwidth]{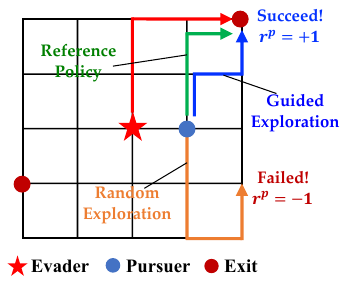}
\caption{Illustration of HMP.}
\label{fig:hmp}
\Description{This figure graphically shows the principle of the heuristic-guided multi-task pre-training (HMP).}
\end{figure}

However, we found that simply applying the MAPPO under the multi-task learning framework can result in low efficiency due to random exploration in the environment. To clarify, consider the example illustrated in Figure~\ref{fig:hmp}, which shows the need for a more efficient pre-training method. Assume that the evader's policy is to take the shortest path to one of the exits (denoted by the red path). If the pursuer explores the environment randomly (the orange path), it will probably lose the game and then receive a negative reward. This situation can occur frequently at the beginning of the pre-training process because the pursuer's initial policy is invariably random. To mitigate this exploration inefficiency\footnote{Notice that many exploration methods in RL such as RND~\cite{burda2018exploration} typically encourage the policy to explore novel states of the environment, which are different from our design where random exploration is less favored.}, we propose a novel scheme: heuristic-guided multi-task pre-training (HMP).

Note that in the RL tasks used for pre-training, we can acquire the evader's policy, which can be used to guide the exploration of the pursuer's policy. Specifically, given the exit node chosen by the evader's policy $\pi^e$, we first induce a reference policy $\hat{\pi}^p$ (represented by the green path) for the pursuer using heuristic methods such as the Dijkstra algorithm. Then, apart from the actions sampled by the generated policy $\pi_{\bm{\theta}}^p$ (Line 6), we also sample the reference actions using the reference policy $\hat{\pi}^p$ and add the data to the training buffer (Lines 8-9). Let $L(\bm{\theta})$ denote the original loss function for training the actor in the MAPPO algorithm. The HMP is implemented by introducing an additional loss into the original loss function: $L = L(\bm{\theta}) + \alpha \text{KL}(\pi^p \Vert \hat{\pi}^p)$ where $\alpha \in [0, 1]$ controls the weight of the guidance of the reference policy and $\text{KL}$ represents the Kullback–Leibler divergence (for the reference policy $\hat{\pi}^p$, the action probability distribution is obtained by setting the probability of the reference action to 1 while all others to 0). 

\subsubsection{\textbf{Stage III: Fine-tuning.}} In this phase, we integrate the pre-trained pursuer policy into the PSRO framework, as shown in Algorithm~\ref{alg:fine-tuning}. The pursuer's policy $\pi_0^p$ is initialized using the output neural network from the pre-trained Grasper, which takes the graphical representation of the specific PEG as an input. Simultaneously, the evader's policy $\pi_0^e$ is randomly initialized (Line 2). Then we follow the standard PSRO framework: in each iteration $k$, the best response (BR) policies for both players, $\pi_k^p$ and $\pi_k^e$, are computed using their respective BR oracles (Lines 4-6). These BR policies are then added to the policy sets $\Pi_k^p$ and $\Pi_k^e$ (Line 7), and the meta-game matrix $U_k$ is updated through simulation (Line 8). Finally, the meta distribution $(\sigma_k^p, \sigma_k^e)$ is computed using any meta-solver (Line 9). 

The BR oracles for both players are the important components of the PSRO algorithm. The evader's BR oracle follows the standard PSRO algorithm given in Algorithm~\ref{alg:psro}. The key difference between our fine-tuning process and the standard PSRO algorithm lies in the training of the pursuer's BR policy, which is highlighted in blue. Specifically, given the pre-trained policy $\pi_0^p$ conditional to the initial conditions, we can use it as the starting point for the computation of the pursuer's BR policy (Line 5), rather than training from scratch. This allows us to simply fine-tune the pre-trained policy $\pi_0^p$ over a few episodes (Line 6) to quickly obtain the BR policy, significantly enhancing the learning efficiency. 

\begin{algorithm}
    \caption{Fine-tuning}
    \label{alg:fine-tuning}
    \textbf{Require}: Grasper, PEG $\mathcal{G}$\;
    $\Pi_0^p=\{\textcolor{blue}{\pi_0^p \gets \text{Grasper}(\mathcal{G})}\}$, $\Pi_0^e=\{\pi_0^e\}$, $U_0$, $\sigma_0^p$, $\sigma_0^e$\;
    \For{$\text{epoch k} = 1, 2, \cdots K$}{
        Compute the evader's BR policy $\pi_k^e$ against $\sigma_{k-1}^p$\;
        \textcolor{blue}{Initialize the pursuer's BR policy $\pi_k^p \gets \pi_0^p$}\;
        \textcolor{blue}{Train $\pi_k^p$ against $\sigma_{k-1}^e$ for few episodes}\;
        Expansion: $\Pi_k^p = \Pi_{k-1}^p \cup \{\pi_k^p\}$, $\Pi_k^e = \Pi_{k-1}^e \cup \{\pi_k^e\}$\;
        Update meta-game matrix $U_k$ through simulation\;
        Compute $\sigma_k^p$ and $\sigma_k^e$ using a meta-solver on $U_k$\;
    }
    \textbf{Return}: $\Pi_K^p$, $\Pi_K^e$, $\sigma_K^p$, $\sigma_K^e$
\end{algorithm}

\begin{figure*}[ht]
\centering
\subfigure[\textbf{Grid Map}]{
\includegraphics[width=0.47\textwidth]{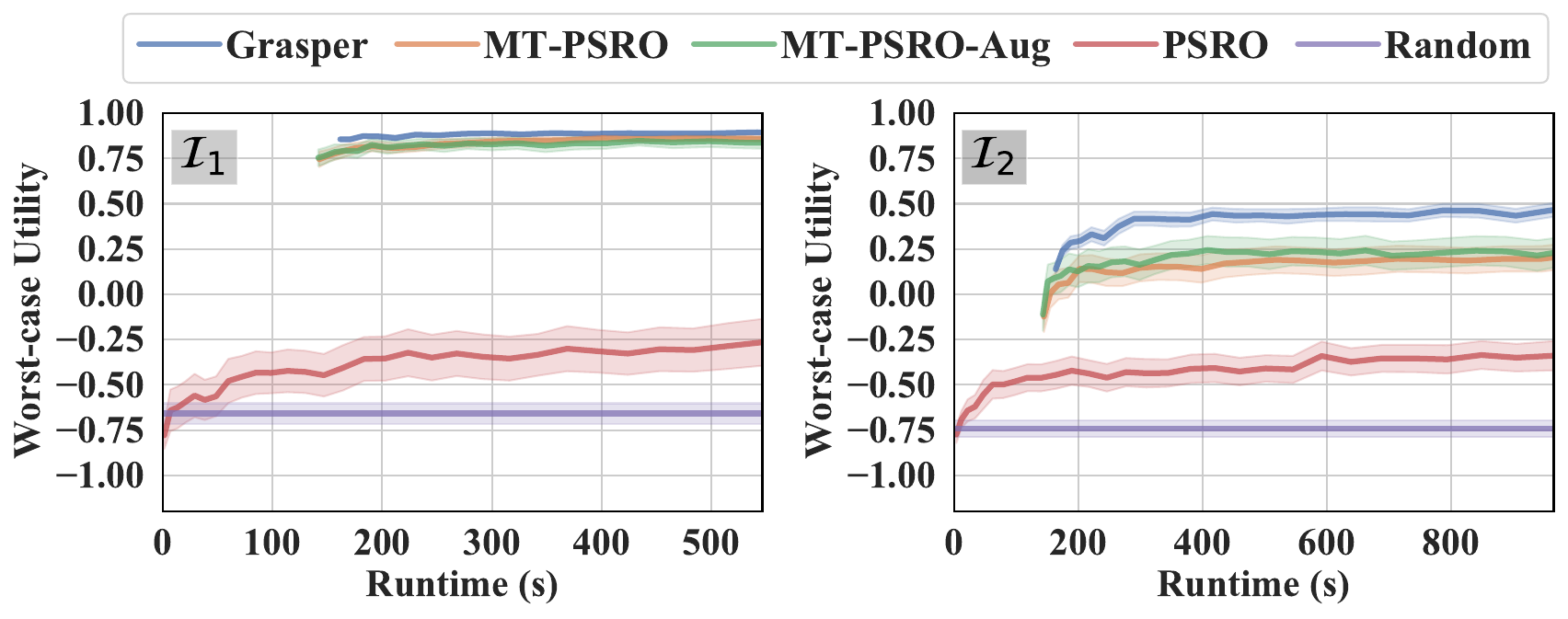}
}\hspace{0.5cm}
\subfigure[\textbf{Scale-Free Map}]{
\includegraphics[width=0.47\textwidth]{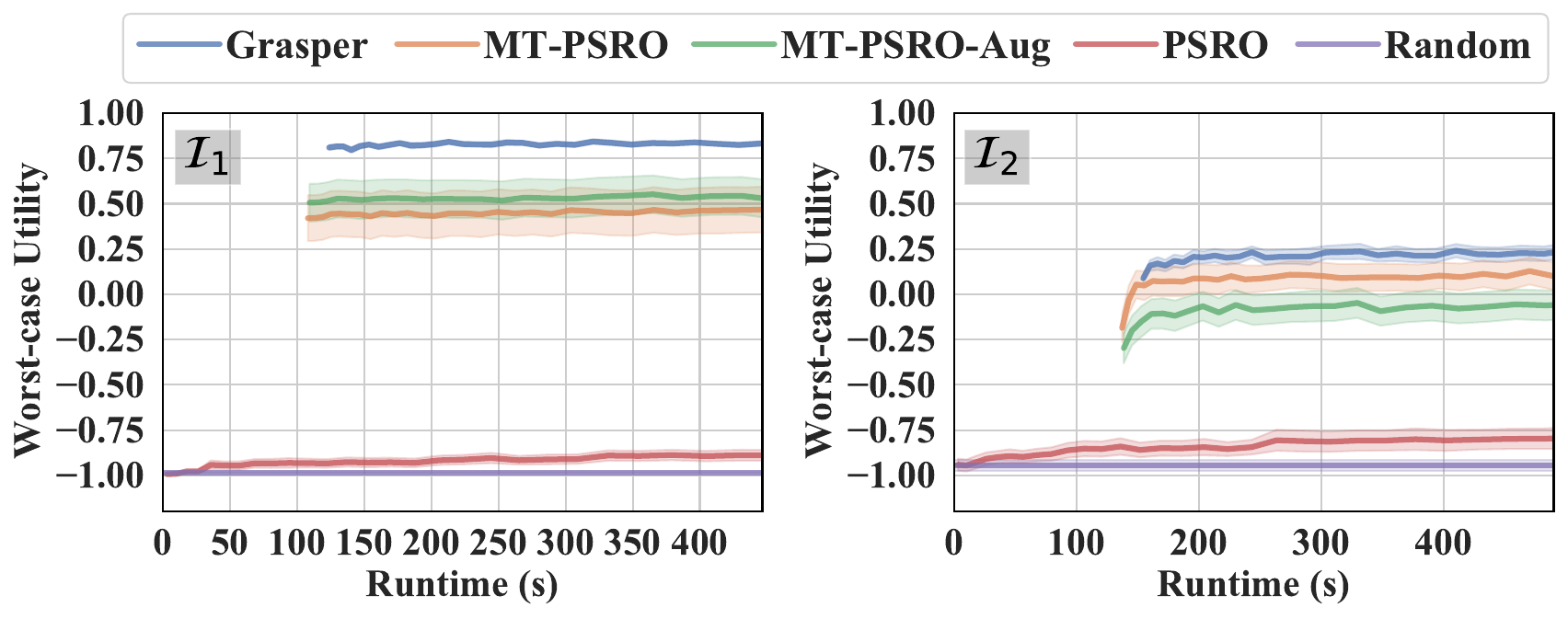}
}
\subfigure[\textbf{Singapore Map}]{
\includegraphics[width=0.47\textwidth]{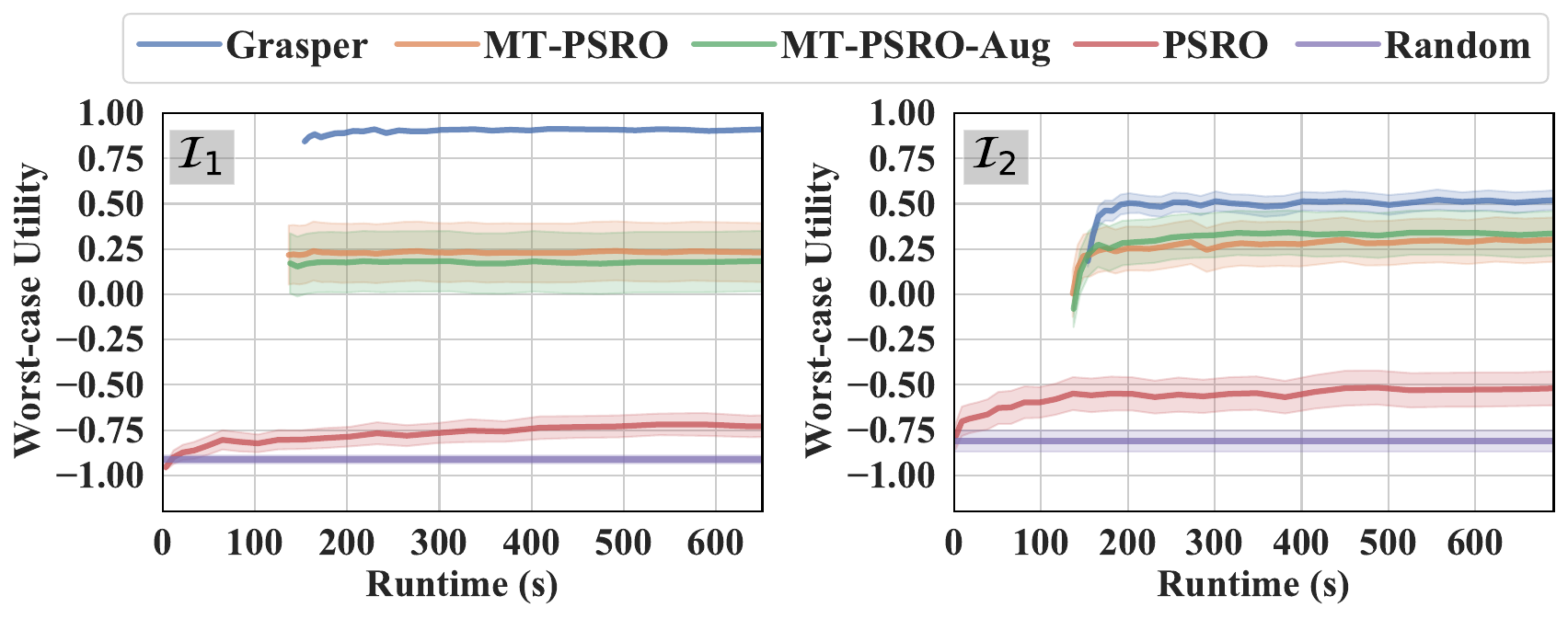}
}\hspace{0.5cm}
\subfigure[\textbf{Scotland-Yard Map}]{
\includegraphics[width=0.47\textwidth]{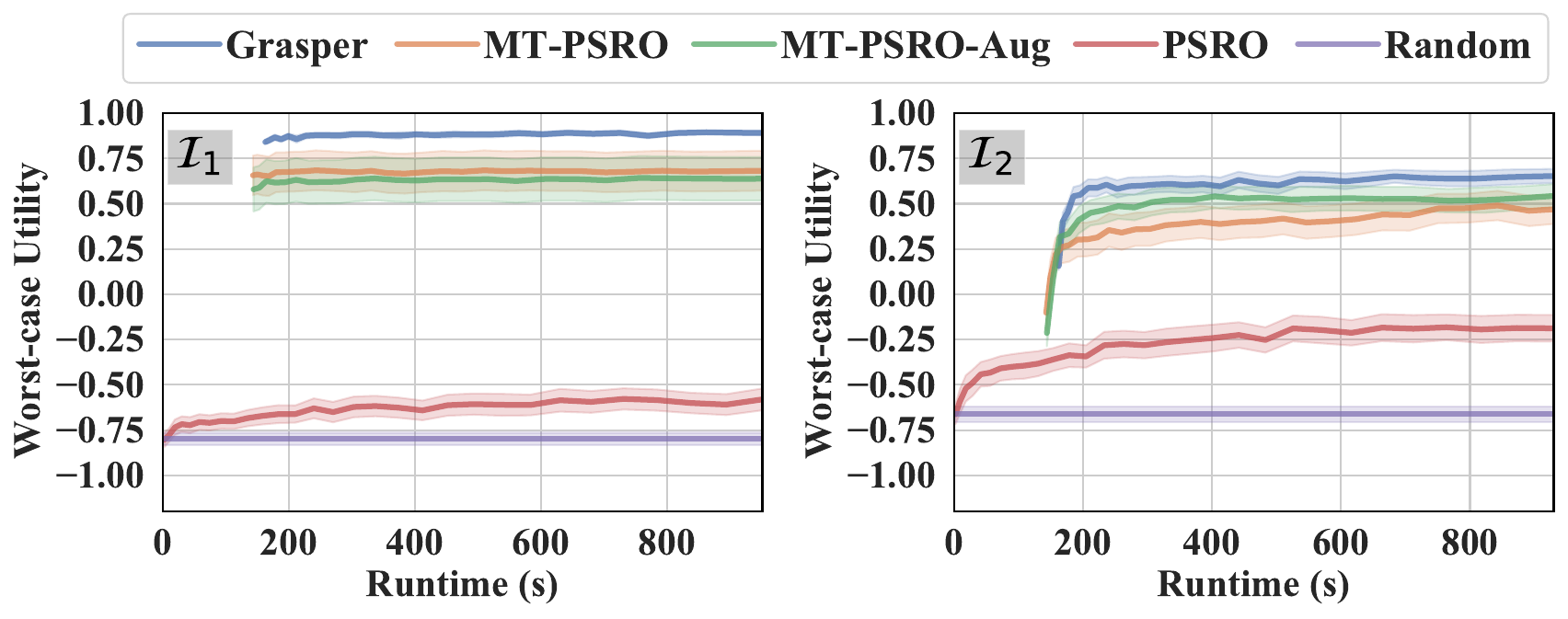}
}
\caption{Evaluation performance. The shaded area represents the standard error.}
\label{fig:utility_vs_epoch_5}
\Description{This figure shows the experimental results in four maps: (a) grid map, (b) scale-free map, (c) Singapore road map, and (d) Scotland-Yard map.}
\end{figure*}

\section{Experiments}

In this section, we perform experiments to evaluate the performance of Grasper and the effectiveness of different components\footnote{The code is available at https://github.com/IpadLi/Grasper.}.

\subsection{Setup}

\textbf{Hyperparameters.} The number of pursuers is $n=5$, the number of exit nodes is 8, the time horizon $T$ is $6 \leq T \leq 10$, and the number of pre-training episodes is 20 million (20M). For PSRO, the number of episodes used for training the best response is 10. We conduct experiments on four maps: the grid map with size $10\times10$, the scale-free graph with 300 nodes, the Singapore map~\cite{xue2021solving} with 372 nodes, and the Scotland-Yard map~\cite{schmid2023student} with 200 nodes. To simulate the situation where a road might be temporally blocked due to congestion or traffic accidents, we set the probability of an edge between two nodes to 0.8 for the grid map, 0.9 for the Singapore map, and 1.0 (i.e., no congestion) for the other two maps. More details on the hyperparameters can be found in Appendix B.

\noindent\textbf{Worst-case Utility.} Given that a PEG is a zero-sum game, we use the pursuer's worst-case utility (as the evader always chooses the shortest path from the initial location to the chosen exit) to measure the quality of the solution: $u^p = \mathbb{E}_{\pi^p \sim \sigma^p, \pi^e \sim \sigma^e} \mathbb{E}[r^p]$, where the inner expectation is taken over the trajectories induced by $\pi^p$ and $\pi^e$ which are respectively sampled according to $\sigma^p$ and $\sigma^e$. 

\noindent\textbf{Training and Test Sets.} (1) We generate $\vert \mathcal{I} \vert = 1000$ games as the training set. During the generation, the minimum length of the evader's shortest path is set to 6 for the grid map and 5 for other maps. (2) We create two test sets, $\mathcal{I}_1$ and $\mathcal{I}_2$, each containing 30 games. (i) $\mathcal{I}_1$ includes the games sampled from the training set $\mathcal{I}_1 \subset \mathcal{I}$ (in-distribution test). (ii) $\mathcal{I}_2$ contains the games distinct from the training set $\mathcal{I}_2 \cap \mathcal{I} = \emptyset$ (out-of-distribution test). To avoid trivial cases (the games that are either too difficult or too simple for the pursuers), we constrain the zero-shot performance of Grasper (i.e., the worst-case utility of the generated policy without fine-tuning) within the range: $[0.8, 0.9]$ for $\mathcal{I}_1$ and $[0.1, 0.2]$ for $\mathcal{I}_2$.

\noindent\textbf{Baselines.} (i) Multi-task PSRO (MT-PSRO): the state-of-the-art (SOTA) approach adapted from~\cite{li2023solving}, which also uses the observation representation layer and HMP. (ii) MT-PSRO with augmentation (MT-PSRO-Aug): the hidden vector obtained from the pre-trained GNN and the time horizon are concatenated to the output of the observation representation layer. (iii) PSRO: the standard PSRO method. (iv) Random: the pursuer randomly selects actions.

\subsection{Results}

The experimental results are summarized in Figure~\ref{fig:utility_vs_epoch_5}. The $x$-axis is the running time. For the purpose of a fair comparison, apart from the running time of the fine-tuning stage (the PSRO procedure), we also include the running time of pre-pretraining and pre-training (called the pre-training time for convenience). Since the games in the training set are uniformly randomly sampled during pre-training, we approximate the pre-training time of each game by averaging the total pre-training time over the training set. Then, for each testing game, we add the pre-training time to the running time (the horizontal gap between 0 and the start of the line). Note that the amortized pre-training time for Grasper, MT-PSRO, and MT-PSRO-Aug is similar since the \textit{pre-pretraining time} is very short (Table~\ref{tab:runtime}). From the results, we can draw several conclusions.

(i) Given a fixed number of episodes for the fine-tuning process, Grasper can start from and converge to a higher average worst-case utility than the baselines, although it takes a certain pre-training time, demonstrating the effectiveness of the pre-pretraining and pre-training in accelerating the PSRO procedure. Note that MT-PSRO and MT-PSRO-Aug also employ pre-pretraining or pre-training, but they perform worse than Grasper, showcasing the superiority of Grasper. (ii) For a fair comparison, MT-PSRO-Aug also integrates the information about the initial conditions of the PEGs. The results clearly show the necessity of the hypernetwork in Grasper. This can be also partly verified by comparing MT-PSRO and MT-PSRO-Aug where their performance is comparable, meaning that naively integrating the information about the initial conditions does not bring much benefit and novel designs are necessary. (iii) An interesting result is that even on the test set $\mathcal{I}_1$ (in-distribution test), MT-PSRO and MT-PSRO-Aug, the strongest baselines, perform worse on all the other maps than on the grid map. We hypothesize the reason is that the other maps are more heterogeneous than the grid map. For example, the degree of the nodes varies from 1 to 16 in the Singapore map while it remains between 2 to 4 in the grid map. Thus, the games generated on the Singapore map share much less similarity. In this sense, the information about the initial conditions is particularly important when solving different PEGs. (iv) In all cases, the performance of Grasper is much more stable than the baselines (smaller standard error) as Grasper can generate distinct policies for different PEGs. In contrast, other baselines either entirely ignore or naively integrate the information about the initial conditions of the PEGs, which renders them hard to generalize to different PEGs, leading to larger performance variance than Grasper. (v) The results on the test set $\mathcal{I}_2$ show that Grasper can solve unseen games, exhibiting better generalizability than the baselines.

\subsection{Ablations}

\noindent\textbf{Effectiveness of Different Modules.} First, we study the contribution of HMP and the observation representation layer (Rep.) to the performance of Grasper, as shown in Table~\ref{tab:ablation-results}. The results show that we can get better performance (high worst-case utility and small standard error) only when combining the two components, meaning that both two components are indispensable for Grasper.

\begin{table}[ht]
\caption{Ablation studies. The results are obtained in the grid map. $\checkmark$ means the module is used.}
\label{tab:ablation-results}
\centering
\setlength{\tabcolsep}{2.7pt}
\begin{tabular}{cc|cccc}
\toprule
\multirow{3}{*}{$\mathcal{I}_{1}$} & HMP &\checkmark &\checkmark &\\
& Rep.& \checkmark & & \checkmark \\
\cmidrule(l){2-6}
& Utility & $\bm{0.90\pm0.01}$ & $-0.54\pm0.06$ & $-0.05\pm0.17$ & $-0.52\pm0.08$ \\
\midrule
\multirow{3}{*}{$\mathcal{I}_{2}$} & HMP &\checkmark &\checkmark &\\
& Rep.& \checkmark & & \checkmark\\
\cmidrule(l){2-6}
& Utility & $\bm{0.45\pm0.04}$ & $-0.60\pm0.06$ & $-0.64\pm0.11$ & $-0.63\pm0.06$\\
\bottomrule
\end{tabular}
\end{table}

\begin{figure}[ht]
\centering
\includegraphics[width=0.94\linewidth]{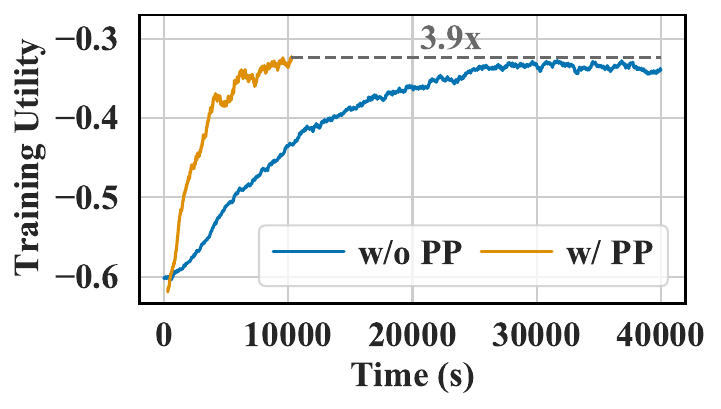}
\caption{Pre-training curves.}
\label{fig:pre-train-curve}
\Description{This figure shows the comparison of pre-training with and without pre-pretraining process.}
\end{figure}

\noindent\textbf{Effectiveness of Pre-pretraining.} Next, we investigate the effectiveness of the pre-pretraining stage in accelerating the whole training procedure of Grasper. Since jointly training GNN and the other parts of Grasper for 20M pre-training episodes requires a long running time, in this ablation study, we focus on the first 2M pre-training episodes and compare the running time of Grasper with pre-pretraining (w/ PP) and without pre-pretraining (w/o PP). The training curves are shown in Figure~\ref{fig:pre-train-curve}, which shows that using pre-pretraining can significantly accelerate the training procedure ($3.9$ times faster than without using pre-pretraining).

\balance

The quantitative values of the running time of the pre-pretraining and pre-training are given in Table~\ref{tab:runtime}. As the pre-pretraining time (304.2 seconds) is much shorter than the pre-training time (9954.9 seconds), the curves of Grasper, MT-PSRO, and MT-PSRO-Aug shown in Figure~\ref{fig:utility_vs_epoch_5} start from a similar position in the $x$-axis.

\begin{table}[hbt!]
    \caption{Running time (second).}
    \label{tab:runtime}
    \centering
    \begin{tabular}{l |l| l}
        \toprule
        & w/o PP & w/ PP \\
        \midrule
        Pre-pretraining & N/A & 304.2 \\
        % \midrule
        Pre-training & 39977.3 & 9954.9  \\
        \midrule
        Total & 39977.3 & 10259.1 (\textbf{3.9x}) \\ 
        \bottomrule
    \end{tabular}
\end{table}

\begin{figure}[ht]
    \centering
    \includegraphics[width=\linewidth]{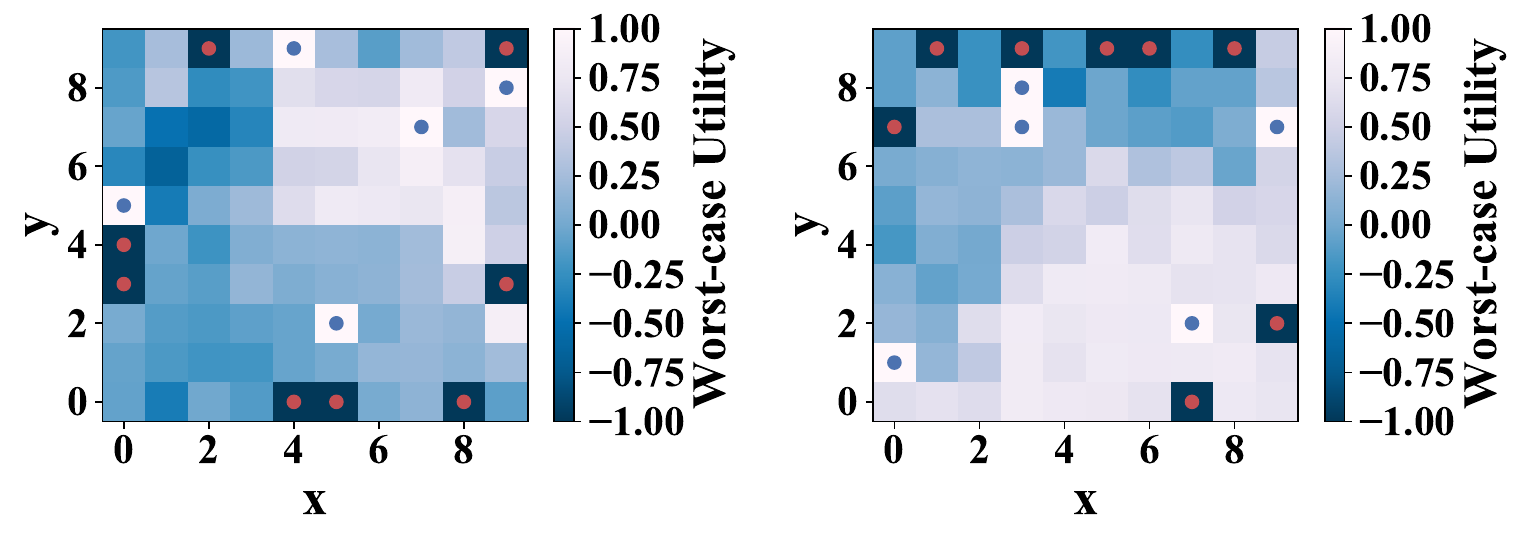}
    \caption{Zero-shot worst-case utility of the pursuer for each possible evader's initial location. Red dots are exits and blue dots are pursuers' initial locations.}
    \label{fig:zero_shot_test_heatmap_5}
    \Description{This figure shows the influence of the evader's initial locations.}
\end{figure}

\noindent\textbf{Influence of Evader's Initial Location.} We perform some experiments using Grasper to provide some insights into the PEG. In Figure~\ref{fig:zero_shot_test_heatmap_5}, we present the pursuer's utility when the evader randomizes the initial location over the grid map. We found that in some areas the pursuers can have high utility. For example, in the top-right of the left figure, there are three pursuers and only one exit, which means it could be hard for the evader to escape. In the bottom-right of the right figure, as the pursuer's initial location is near the two exits, it could be easy for the pursuer to catch the evader, even though there is only one pursuer in this area. The results reflect the intuition that Grasper can generate distinct policies for different games and hence, the performance is more stable.

\section{Conclusions}

In this work, we investigate how to efficiently solve different PEGs with varying initial conditions. First, we propose a novel generalizable framework, Grasper, which includes several critical components: (i) a GNN to encode a specific PEG into a hidden vector, (ii) a hypernetwork to generate the base policies for the pursuers conditional on the hidden vector and time horizon, (iii) an observation representation layer to encode the pursuers' observations into compact and meaningful representations. Second, we introduce an efficient three-stage training method which includes: (i) a pre-pretraining stage that learns robust PEG representations through GraphMAE, (ii) a heuristic-guided multi-task pre-training stage that leverages a reference policy derived from heuristic methods such as Dijkstra to regularize pursuer policies, and (iii) a fine-tuning stage that utilizes PSRO to generate pursuer policies on designated PEGs. Finally, extensive experiments demonstrate the superiority of Grasper over baselines in terms of solution quality and generalizability. To the best of our knowledge, this is the first attempt to consider the generalization problem in the domain of PEGs. Future directions include (i) more efficient task sampling strategies for pre-training, e.g., AdA~\cite{team2023human}, (ii) a model capable of generalizing to different PEGs with different underlying graph topologies, e.g., generalizing from grid maps to scale-free maps, and (iii) a model capable of tackling more complex settings, e.g., learning-based evader.

%%%%%%%%%%%%%%%%%%%%%%%%%%%%%%%%%%%%%%%%%%%%%%%%%%%%%%%%%%%%%%%%%%%%%%%%

%%% The next two lines define, first, the bibliography style to be 
%%% applied, and, second, the bibliography file to be used.

\begin{acks}
This work is supported by the National Research Foundation, Singapore under its Industry Alignment Fund -- Pre-positioning (IAF-PP) Funding Initiative. Any opinions, findings and conclusions, or recommendations expressed in this material are those of the author(s) and do not reflect the views of National Research Foundation, Singapore. Youzhi Zhang is supported by the InnoHK Fund. Hau Chan is supported by the National Institute of General Medical Sciences of the National Institutes of Health [P20GM130461], the Rural Drug Addiction Research Center at the University of Nebraska-Lincoln, and the National Science Foundation under grant IIS:RI \#2302999. The content is solely the responsibility of the authors and does not necessarily represent the official views of the funding agencies.
\end{acks}

\bibliographystyle{ACM-Reference-Format} 
\bibliography{sample}

%%%%%%%%%%%%%%%%%%%%%%%%%%%%%%%%%%%%%%%%%%%%%%%%%%%%%%%%%%%%%%%%%%%%%%%%

\end{document}